%% file: main.tex
\documentclass[runningheads]{llncs}
\usepackage[year=2024,ID=11710]{eccv}
\usepackage{eccvabbrv}
\usepackage{graphicx}
\usepackage{booktabs}
\usepackage{svg}
\usepackage[accsupp]{axessibility}  
\usepackage{xcolor}

\usepackage{subcaption}

\usepackage[pagebackref,breaklinks,colorlinks,citecolor=eccvblue]{hyperref}

\usepackage{orcidlink}

\def\CName{CAMOUFLaGE}
\def\BName{\CName{}-Base}
\def\IName{\CName{}-Light}

\begin{document}

\title{Latent Diffusion Models for Attribute-Preserving Image Anonymization}

\author{Luca Piano\inst{1}\orcidlink{0000-0003-4467-7358} \and
Pietro Basci\inst{1}\and
Fabrizio Lamberti\inst{1}\orcidlink{0000-0001-7703-1372} \and
Lia Morra\inst{1}\orcidlink{0000-0003-2122-7178}}

\authorrunning{Piano et al.}

\institute{Politecnico di Torino, Turin, Italy \\ \email{name.surname@polito.it}}

\maketitle

\input{sec/0_abstract_final}    
\input{sec/1_intro_new}
\input{sec/2_rwork}
\input{sec/3_method}
\input{sec/4_expsettings}
\input{sec/5_results}
\input{sec/6_conclusion}
\input{sec/7_ack}

%
%
\bibliographystyle{splncs04}
\bibliography{main}

\end{document}

%% file: sec/0_abstract_final.tex
\begin{abstract}
Generative techniques for image anonymization have great potential to generate datasets that protect the privacy of those depicted in the images, while achieving high data fidelity and utility. Existing methods have focused extensively on preserving facial attributes, but failed to embrace a more comprehensive perspective that considers the scene and background into the anonymization process. This paper presents, to the best of our knowledge, the first approach to image anonymization based on Latent Diffusion Models (LDMs). Every element of a scene is maintained to convey the same meaning, yet manipulated in a way that makes re-identification difficult. We propose two LDMs for this purpose: CAMOUFLaGE-Base exploits a combination of pre-trained ControlNets, and a new controlling mechanism designed to increase the distance between the real and anonymized images. CAMOFULaGE-Light is based on the Adapter technique, coupled with an encoding designed to efficiently represent the attributes of different persons in a scene.  The former solution achieves superior performance on most metrics and benchmarks, while the latter cuts the inference time in half at the cost of fine-tuning a lightweight module. We show through extensive experimental comparison that the proposed method is competitive with the state-of-the-art concerning identity obfuscation whilst better preserving the original content of the image and tackling unresolved challenges that current solutions fail to address. 
\end{abstract}

%% file: sec/1_intro_new.tex
\section{Introduction}
\label{sec:intro}

The widespread availability of personal photos and videos shared on social media platforms is raising increasing concerns about individual privacy\cite{shoshitaishvili2015portrait}. The potential misuse of such data without the owner's consent has led to stricter legislation, notably the General Data Protection Regulations (GDPR) in the European Union. GDPR mandates explicit consent from data owners for any use of their personal information. While this is a significant step towards safeguarding data privacy and ownership, it poses challenges for researchers seeking to assemble high-quality datasets that include images containing people. In response to these concerns, a growing area of research devoted to image anonymization has emerged. The objective is to develop solutions that render subjects unidentifiable, removing sensitive information, and thus making the images freely usable. 

\begin{figure}[tb]
   \centering
   \includegraphics[width=\linewidth]{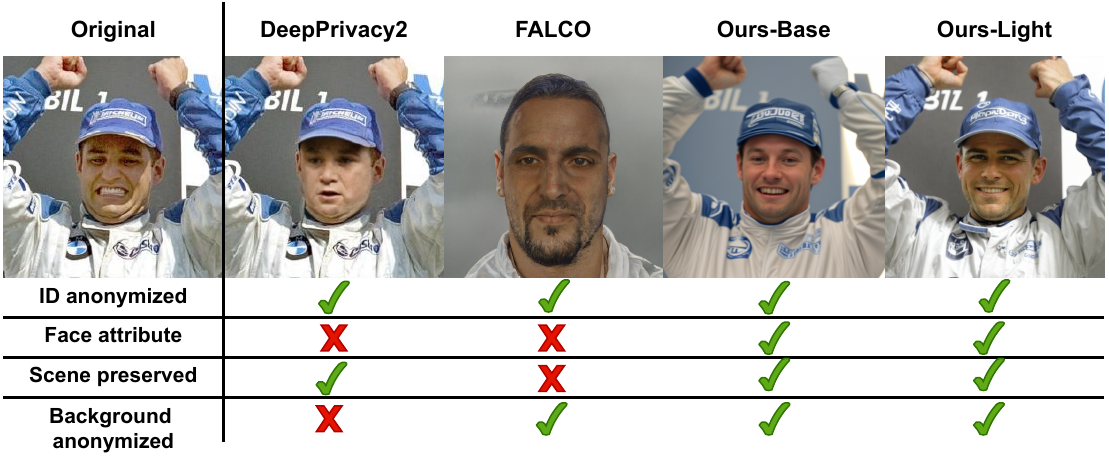}
   \caption{Comparison of the proposed method to DeepPrivacy2\cite{hukkelaas2023deepprivacy2} and FALCO\cite{barattin2023attribute} in terms of identity anonymization, face attribute preservation, scene preservation and background anonymization.}
   \label{fig:intro}
\end{figure}

Early methods such as \textit{pixelation}, \textit{blurring}, or \textit{masking} distort the images to an extent that makes them unusable for downstream applications\cite{barattin2023attribute}. More recent methods exploit Generative techniques such as Generative Adversarial Networks (GANs) with the ultimate goal of generating images that, while anonymous, are still realistic and close to the original distribution. Despite great progresses in the quality and fidelity of the resulting images, key limitations still persist. 
On the one hand, \textit{inpainting-based techniques} such as DeepPrivacy \cite{hukkelaas2019deepprivacy} and DeepPrivacy2 \cite{hukkelaas2023deepprivacy2} only change sensitive information (the face, or at most the body): leaving the background intact poses an evident privacy risk as one can potentially re-identify the original image from the background itself \cite{barattin2023attribute}. On the other hand, generative methods such as FALCO \cite{barattin2023attribute} recreate completely synthetic images that preserve some attributes of the original image (namely, facial attributes), while completely changing the background, clothes, accessories, gestures, and so forth. This approach achieves strong anonymization metrics, at the risk of completely changing the image composition, plastic features and cultural connotations \cite{santangelo2023}. Indeed, we argue that an anonymization technique should not merely strive to preserve the same facial attributes of the original image. To preserve the same cultural meaning, specific objects or elements that could be used to identify a person (such as a distinctive hairstyle, a tattoo, a particular choice of clothing, or an object in the scene) should not be removed altogether, but manipulated  in a way that makes re-identification  difficult, as shown in \cref{fig:intro}.

\textit{In an effort to strike a balance between privacy and fidelity, we propose} \textit{\CName{}},  \textit{a new anonymization framework based on Latent Diffusion Models} (LDMs).  Our approach surpasses the limitations of inpainting-based strategies by recreating the entire image, so that an attacker could not resort to the background for re-identification. Simultaneously, important features of the background and scene composition, as well as key facial attributes preserved, but subtle mutations are introduced to enhance privacy. This approach yields greater data fidelity as well as data utility, as more  properties of the original images are preserved for downstream tasks \cite{jordon2022synthetic}. 

The main contributions of this paper can be summarized as follows:
\begin{itemize}
    \item To the best of our knowledge, we are the first to exploit LDMs, specifically Stable Diffusion \cite{rombach2022high}, for anonymization purposes, proposing two architectures  \BName{} and \IName{}.
    \item We experimentally show that \CName{} lowers the risk of re-identification due to the background compared to inpainting-based techniques.  
    \item Compared to state-of-the-art methods, \CName{} better reproduces a wide range of features of the original image,  demonstrated by qualitative and quantitative assessment,even in complex scenes with multiple persons.
\end{itemize}

%% file: sec/2_rwork.tex
\section{Related work}
\label{sec:rwork}
\subsection{Image anonymization}
 
Traditional \textit{obfuscation-based anonymization} techniques include pixelation, blurring, and masking \cite{boyle2000effects, tansuriyavong2001privacy,chen2007tools}.
Despite their widespread use, these methods introduce distortions that render images unable for many downstream applications, such as for instance those that rely on face detection or facial attributes. In addition, obfuscation-based techniques such as blurring are vulnerable to reversion \cite{gross2006model,neustaedter2006blur}.  With the advent of GANs, \textit{conditional generative models} have become the state-of-the-art method for image anonymization  \cite{hukkelaas2019deepprivacy, hukkelaas2023deepprivacy2, maximov2020ciagan, barattin2023attribute, boyle2000effects, chen2007tools, tansuriyavong2001privacy, wen2022identitydp, wu2018privacy, shan2020fawkes, li2019anonymousnet}. They strive to achieve more robust anonymization, while maintain key properties of the original images; consequently, the resulting anonymized images remain suitable for a wider range of downstream tasks.

DeepPrivacy \cite{hukkelaas2019deepprivacy} is based on a  GAN \cite{mirza2014conditional}, conditioned on the original pose and background.
However, it is limited in handling irregular poses, complex backgrounds, and high occlusions, leading to potential face corruption.
Hukkelås et al. extended the tool to full-body in DeepPrivacy2 (DP2) \cite{hukkelaas2023deepprivacy2} by employing a detection module and three generators for faces and bodies.
While this approach enhances both image quality and identity obfuscations, certain limitations persist. Notably, it generates a restricted range of identities based on a given input condition, and it is unable to guarantee complete anonymization due to the potential failure of the detector. 

CIAGAN \cite{maximov2020ciagan} anonymizes faces and bodies using an  inpainting GAN conditioned on several inputs, including the image, a landmark-based representation of the face, a masked face, and the desired identity.
Anonymization is achieved through an identity swap performed as an image inpainting task.
Similar to previous approaches, however, its performance depends on the accuracy of the landmark detector.
Additionally, the ID-swapping technique utilizes real identities to guide the generation of the anonymized face, which raises critical issues from a security and privacy standpoint. IdentityDP \cite{wen2022identitydp} also employs a form of identity swapping, using separate identity and attribute encoders.

The closest work to us is FALCO \cite{barattin2023attribute}, a method specifically designed to retain facial features during anonymization.
It optimizes the image representation in the latent space of a pre-trained StyleGAN2\cite{karras2020analyzing} to achieve both anonymization and facial attribute preservation. Given a real dataset \(\mathcal{X}_{R}\), a randomly generated fake dataset \(\mathcal{X}_{F}\) is created and the corresponding latent codes in the StyleGAN2\cite{karras2020analyzing} latent space \(\mathcal{W}^{+}\) are extracted for both real and fake faces. To obtain a relevant starting latent code that will be then optimized, the nearest fake latent code in the feature space of the  ViT-based FaRL\cite{zheng2022general} image encoder is selected, thus bypassing the need of using real persons to perform identity swapping.  It is then optimized based on a loss designed to increase the distance of real and fake faces according to the ArcFace\cite{deng2019arcface} identity representations. 
While FALCO demonstrate good qualitative and quantitative performance, it is still limited to face anonymization, whereas the proposed  method aims at retaining key features of all persons, as well as the overall scene and background.

\subsection{Conditional image synthesis}

In the landscape of high-resolution image synthesis, the comparison between GANs \cite{goodfellow2014generative} and Diffusion Models (DMs)\cite{rombach2022high, ramesh2022hierarchical, balaji2022ediffi, nichol2021glide} is crucial for understanding recent advancements in the field. GANs, although initially promising for their ability to generate realistic images, face limitations in scaling to complex, multi-modal distributions, especially in scenarios with high variability. DMs not only overcome training instabilities and mode-collapse associated with GANs, but also offer a general-purpose conditioning mechanism for multi-modal tasks. In fact, by decomposing the generation process in a sequence of denoising steps, they allow for guiding mechanisms to steer the image generation toward the desired content without retraining the model. More specifically, LDMs are class of DMs designed to scale to high resolution images by operating in a perceptually equivalent but computationally efficient latent space computed by an autoencoder  \cite{saharia2022photorealistic}.

Many existing LDMs, such as the widely adopted Stable Diffusion (SD) model \cite{rombach2022high}, are predominantly \textit{text-to-image models}, conditioning the output on textual features \cite{radford2021learning}. However, designing effective text prompts for such models can be challenging, often necessitating intricate prompt engineering \cite{gal2022image}. Text prompts also offer limited control and expressivity for complex scenes \cite{ye2023ip,zhang2023adding}. Recognizing these limitations, various studies have explored alternative forms of conditioning.
Some studies, exemplified by SD Image Variations \cite{SDIV} and Stable unCLIP \cite{stable-diffusion-2-1-unclip}, have showcased the efficacy of fine-tuning text-conditioned DMs directly on image embeddings to steer the generation process towards a specific content. More recently, techniques like ControlNet \cite{zhang2023adding}, T2I-adapter \cite{mou2023t2iadapter}, and IP-Adapter \cite{ye2023ip} have combined spatial conditioning prompts, such as a sketch or semantic segmentation, with existing text-to-image DMs. In this work, we build on such techniques to tackle the delicate balancing act between content preservation at large and identity obfuscation. 

%% file: sec/3_method.tex
\section{Methodology}
\label{sec:method}

\CName{} tackles the anonymization task via a meticulously conditioned LDM. First, non-sensitive representations are extracted from the original image. Second, the reconstruction phase leverages them to guide the diffusion process to preserve the original content and composition. Two architectures are compared: \BName{} inspired by ControlNet \cite{zhang2023adding}, and \IName{}, a lightweight architecture based on  T2I-Adapter \cite{mou2023t2iadapter}.

\subsection{\BName{} }

\begin{figure*}[tb]
    \centering
    \includegraphics[width=\textwidth]{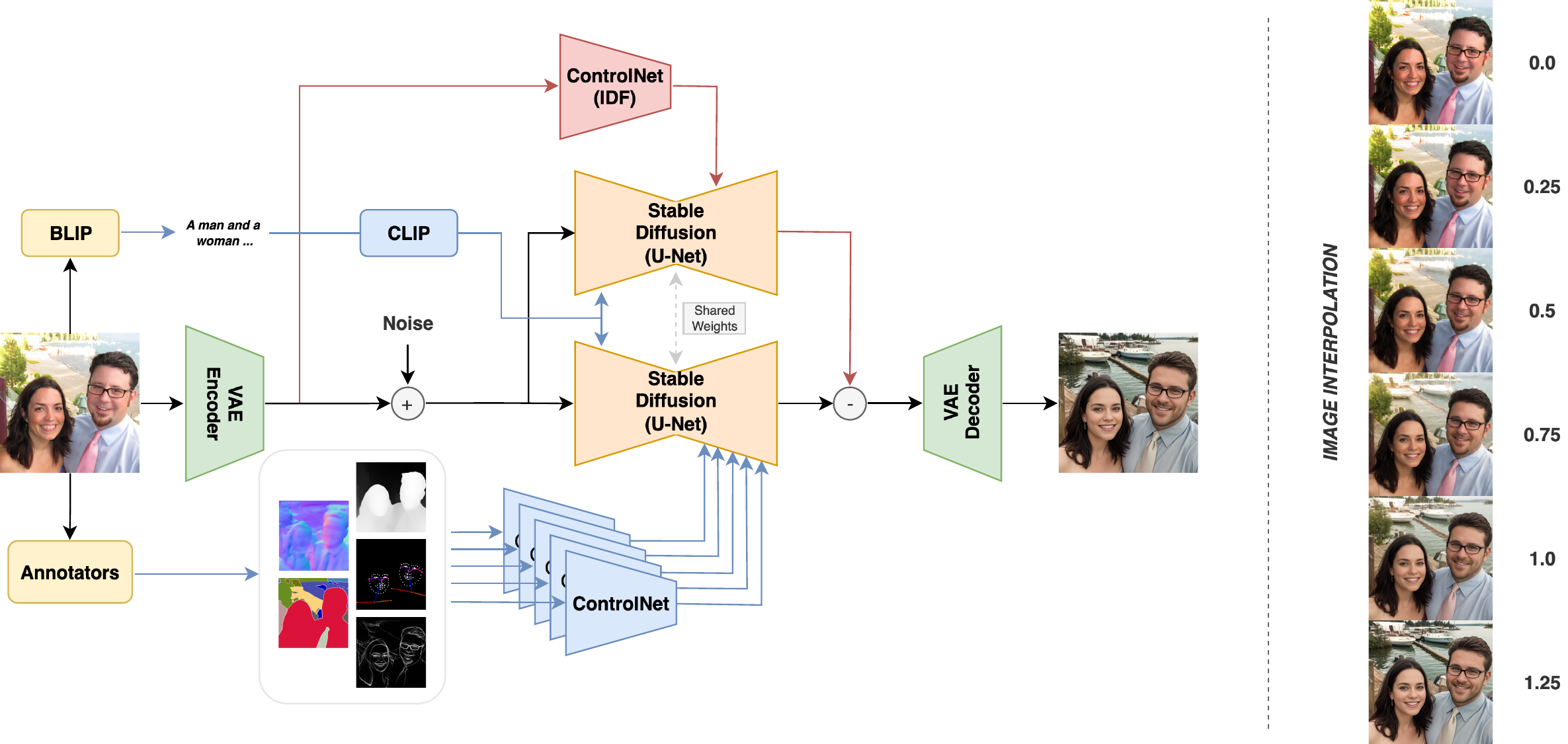}

    \caption{Left: architecture of \BName{}. The \textcolor{orange}{annotators} extract five different types of spatial information and a caption from the input image, which are used to guide Stable Diffusion at inference time via a \textcolor{cyan}{ControlNet} module as \textit{positive control}. The diffusion process starts from the encoding of the original image in the latent space, after adding some noise. An additional ControlNet, taking as input directly the Variational Autoencoder (VAE) encoding and acting as the identity function (\textcolor{purple}{IDF}), is used as \textit{negative control} through a classifier-free guidance-like mechanism, controlled by an \textit{anonymization scale} parameter $a_s$. All components are pretrained and no fine-tuning is needed, except for the IDF component. Right: Anonymized images at varying anonymization scales $a_s$. Between $a_s=0$ and $a_s=1.0$ \CName{} interpolates between the original image and the image synthesized from the positive controls. At $a_s>1.0$ the negative control is enabled further pushing the output away from the original image within the limits afforded by the positive controls.    
   }
    \label{fig:baseline}

\end{figure*}

The \BName{} architecture, shown in \cref{fig:baseline}, is based on multiple ControlNets \cite{zhang2023adding}. ControlNet was designed to take as input a variety of different spatial conditioning prompts, including line sketches, edges, semantic segmentation, surface normals, depth maps, and human pose. In our case, the information is extracted from the image to be anonymized using pretrained models.  An ensemble of multiple controls achieves tighter control over the generated images, while compensating the weaknesses of each individual control and preventing artifacts especially in complex scenes.

\paragraph{Information extraction, spatial and caption conditioning} 
At a preliminary analysis, edge and lineart-based controls yielded a higher re-identification rate, but also consistently produced higher quality images with fewer artifacts and preserved details such as text, glasses, earrings, necklaces, hairstyle, gender, and so forth. For this reason, \BName{} combines five ControlNets taking as input the depth map, surface normal map, segmentation, human pose, and lineart edges. 
The intensity of each control was properly weighted to find an acceptable trade-off between image quality and risk of re-identification: 0.5 (depth), 0.3 (normal), 0.3 (segmentation), 0.4 (pose), 0.5 (lineart). In addition, given that the highest risks of re-identification is associated to the lineart control, it was completely disabled after 50\% of the sampling steps. Cutting off the Lineart control prematurely allows to inject the desired information in the early diffusion steps, in which the overall content of the generated image is defined,  while leaving the model free to introduce smaller variations during the later steps.  

The global representations, used to guide the generation process, are extracted from the original image using pre-trained models and/or image processing software \cite{control_net_repo,cao2017realtime,ranftl2020towards,bae2021estimating,jain2023oneformer} (denoted as Annotators in \cref{fig:baseline}). All experiments were conducted on the pre-trained ControlNet version 1.1 \cite{zhang2023adding}. The \textit{caption} is generated from the original image using BLIP \cite{li2022blip}, projected into a latent representation using the CLIP ViT-L/14 text encoder and then directly injected into the U-Net through cross-attention layers. To remove unwanted features (e.g., artifacts) and improve image quality, we used a negative prompt obtained by training an embedding through textual inversion \cite{gal2022image}. Further details are available in the Supplementary Material.

\paragraph{Anonymization guidance} When reconstructing images using only privacy-safe information, as will be detailed in Section \ref{sec:results}, the risk of re-identification is higher than the comparable literature. In fact, the diffusion process starts from the latent representation of the original image plus some random noise; also depending on the chosen random seed, the reconstructed image may be too similar to the original one.
Exploiting the higher controllability of DMs allowed by the iterative generation process, we introduced an \textit{image based classifier-free guidance} (CFG) \cite{ho2022classifierfree} to combine the positive conditioning provided by the annotators with a negative control mechanism that remove unwanted features during the denoising process, thus lowering the re-identification risk. 
Specifically, the negative control is provided by an additional ControlNet, that takes as input the VAE encoding of the original image and thus acts as an identity function that reconstructs faithfully the original image. 
At each step the latent vector predicted by the identity model is subtracted from the final prediction to steer the anonymized image away from the original one, within the small margin of freedom afforded by the strong positive controls.

Mathematically, if the whole original image is denoted as condition $c_{2}$, all the representations $c_{1}$ extracted from the image encode, to some extent, a subset of the information carried by $c_{2}$ (i.e., $c_{1} \subset c_{2}$). The CFG emphasizes information within $c_{1} \setminus c_{2}$ and attenuates information within $c_{2} \setminus c_{1}$ with a strength provided by a weight $a_s$, while information within $c_{1} \cap c_{2}$ remain unaffected. At each step, the noise prediction is updated using a modified CFG, in which the prediction of the identity model is added to interpolate between the original and the anonymized version:

\begin{equation} \label{eq:enh-cfg1}
    \small
    \hat{\epsilon}_{\theta}(x_{t}|c, p) := \mathbf{(1 - a_s) \: \epsilon_{\theta}(x_{t}|c_{2}, p_{+})} + \mathbf{w_1} \: \epsilon_{\theta}(x_{t}|c_{1}, p_{-}) + \mathbf{w_2} \: \epsilon_{\theta}(x_{t}|c_{1}, p_{+});
\end{equation}
where $\epsilon_{\theta}$ is the conditional denoising autoencoder, $p_{+}$ and $p_{-}$ refer to the prompt (caption) and the negative prompt (caption from which the resulting image must differ), 
respectively, $w_1 = \min (a_s, 1) * (1-\omega)$ is the weight of the negative prompt, $w_2 = a_s * \omega$ if $0 < a_s < 1$, $(a_s - 1 + \omega)$ if $a_s \ge 1$ is the weight of the positive prompt,
and $a_s$ is an adjustable parameter that controls the \textit{anonymization scale}. \cref{fig:baseline} shows an example with $a_s$ ranging from 0.0 to 1.25.

\paragraph{Reconstruction and sampling} The generative model used is Realistic Vision V5.1, which is derived from Stable Diffusion v1.5\cite{stable_dif1_5}, which generates higher quality images reducing artifacts when faces are appearing in complex scenes. The original VAE \cite{kingma2013auto} was replaced with VAE ft-MSE \cite{sd-vae-ft-mse} which slightly improves face quality while keeping the same downsampling factor of 8. To mitigate persisting artifacts on small faces, in particular on details such as eyes and mouth, instead of using a VAE with a lower down-sampling factor, images with higher resolution were generated. Even through SD is a fully convolutional network, selecting a higher output resolution (\eg, 1024 or more) may generate repetitions, such as multiple arms or heads. This issue, extremely evident when conditioning only on a text prompt, 
is significantly reduced if the generation process is conditioned by a fairly strong control, including multiple modalities. The model was tested on images at resolution  \(768^2\) (i.e., $2.25\times$ the training resolution), images with the longest side at 1024 ($\sim3\times$) and images with the shortest side at 1024 ($\sim6\times$).  

The scheduler  DPM++ SDE Karras \cite{karras2022elucidating}, based on preliminary results, achieves the best results in terms of quality and requires a low number of steps (only 16 steps in the experiments). Thanks to the higher control strength, a higher noise (0.9) can be added to the encoded images. 

\subsection{\IName{}}
\begin{figure*}[t]
    \centering
        \includegraphics[width=0.8\linewidth]{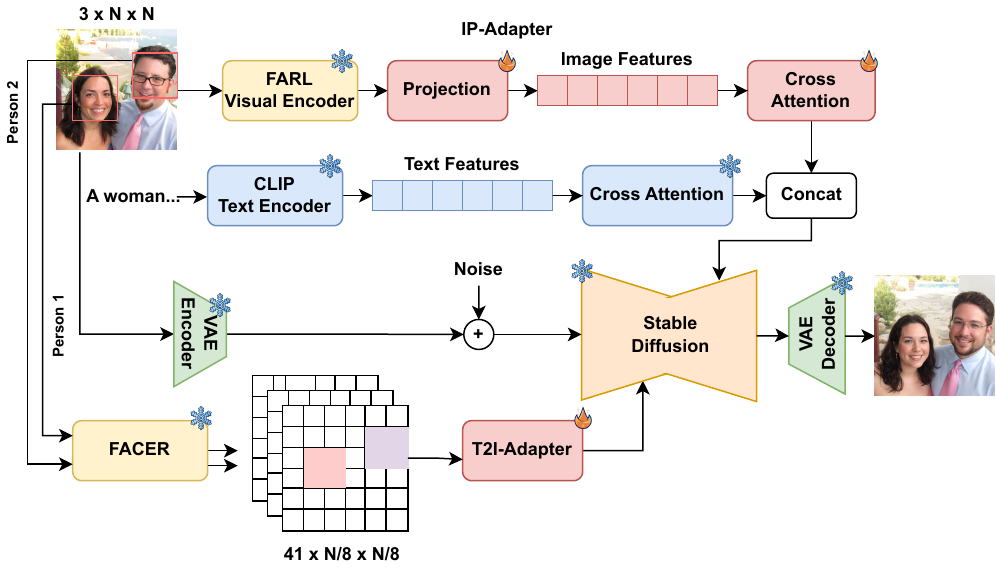}
    \caption{Architecture of  \IName{}. IP-Adapter is fine-tuned to condition the image generation process by means of a decoupled cross-attention layer taking as input an image embedding (extracted from the FaRL visual encoder, to enhance control on visual features) and the caption.  The caption is used only at training time to aid in encoding key scene characteristics. To prevent features from different persons from mixing, an ad-hoc encoding was designed to spatially encode 40 facial attributes, extracted from the FACER pre-trained model, and facial keypoints for each individual person. This encoding is given as input to a T2I-Adapter module, jointly trained with the IP-Adapter module.  During the training, the SD module and the various encoders are kept \textcolor{blue}{frozen}, while the IP-Adapter and T2I-Adapter module are \textcolor{red}{trainable}.  }
    \label{fig:proposed}

\end{figure*}

\BName{} relies on information extracted by annotators that are not immune to hallucination. One example out of all is that of captions: the BLIP model generates accurate descriptions of images most of the time, but occasionally makes errors, which can lead to incorrect generations. In addition, captions do not always guide generation in the desired way (\eg, when a color is used in the caption it may be spread to multiple parts of the image, see Supplementary Material). Moreover, since each control needs a separate ControlNet, the inference time rises significantly. \IName{}, instead, relies on IP-Adapter \cite{ye2023ip}, a lightweight solution for enhancing pre-trained text-to-image diffusion models with image prompts, thus reducing the need for complex prompt engineering. Comprising an image encoder and decoupled cross-attention modules, it efficiently incorporates image features into the model. 
As in \BName{}, the generation process is conditioned from features extracted from the image to be anonymized, as shown in \cref{fig:proposed}. Specifically, \textit{we design a feature extraction module that separately encodes information extracted from each person present in the scene}.

\paragraph{Information extraction and training}
Instead of the ViT-H/14 transformer from CLIP, as in the original IP-Adapter implementation, we chose as encoder the Facial Representation Learning (FaRL) ViT-based 512-dimensional feature space \cite{zheng2022general}. FaRL is a universal facial representation scheme trained in a contrastive way on 20 million face-text pairs.
In preliminary experiments, we noticed that IP-Adapter had a tendency to disregard spatial information and thus generate \textit{images in which multiple subjects shared the same facial features},  such as ethnicity, presence of accessories, and hair color. To overcome this problem, we coupled the IP-Adapter with \textit{T2I-Adapter}, a much lighter version of ControlNet, which takes as input \textit{separate facial features for each person}. From FACER \cite{zheng2022general} we extracted 40 different facial attributes and the position of selected keypoints (eyes, mouth and nose). These attributes are incorporated as 41 separate channels of a feature matrix, with dimensions matching those of the U-Net input, at each position corresponding to a face in the original image.

Training was performed using the IP-Adapter module alone for the first 120k steps, and then continued for 100k steps using both IP-Adapter and T2I-Adapter. In both training phases, the FFHQ-in-the-wild dataset was used as the primary training set. More training details are provided in the Supplementary material. 

\paragraph{Reconstruction, sampling and identity swapping} 
While IP-Adapter preserves substantial information from the original image, it also retains features crucial for potential re-identification. To reduce the re-identification rate, identity swapping was employed at inference time, wherein faces within the images were systematically replaced with comparable alternatives. To avoid exploiting the identities of real persons, a dataset of 30,000 images, similar to the one employed by FALCO \cite{barattin2023attribute}, was generated. Each face was substituted with the most similar one having a minimum distance of 1.0 from features extracted by FARL. This methodology ensured that the replaced face retained characteristic information without facilitating re-identification, as exemplified in Supplementary Material. 
Finally, a noise of 0.6 was added to the resulting encoding and added as input to the U-Net. Inspired by  MixUp \cite{zhang2017mixup} data augmentation, both the original and face-swapped images are given as input to the IP-Adapter module.

The same inference configuration used for the \BName{} was used for image generation with minor differences: generation takes place at a resolution of \(768^2\) pixels with 30 steps, and captions are not used at inference time.

%% file: sec/4_expsettings.tex
\section{Experiments}
\label{sec:exp_settings}

The proposed method was qualitatively and quantitatively evaluated on images from several datasets considering different points of view: image quality and fidelity, re-identification risk, and computational requirements.

\paragraph{Datasets}
We evaluated the quality of the anonymized images on four datasets encompassing a variety of images typical of social media, from portraits to persons in context: (i) Flickr-Faces-HQ (FFHQ) in-the-wild \cite{karras2019style} composed of 70k images extracted from Flickr,  (ii) WIDER FACE \cite{yang2016wider} consisting of 32,203 images selected from the publicly available WIDER dataset, (iii) Labeled Faces in the Wild (LFW) \cite{huang2008labeled} composed of 13,233 images of 5,749 people with a resolution of \(250^2\) pixels, (iv)  CelebA-HQ\cite{karras2017progressive} consisting of 30k images with a resolution of \(1024^2\) pixels.
Anonymization was evaluated on LFW and CelebA-HQ.

\paragraph{State of the art comparison}
The proposed methods were evaluated against two state-of-the-art anonymization frameworks, DP2 and FALCO (with a resolution respectively of \(250^2\) and \(1024^2\)), utilizing the publicly available implementations by the respective authors, while the proposed methods operated at a resolution of \(768^2\) to optimize for reduced inference time. 

\paragraph{Image quality and fidelity}
Visual DNA \cite{ramtoula2023visual} is a recently introduced technique that compares individual images as well as datasets based on the distributions of neuron activations throughout the layers of a pre-trained feature extractor, the Mugs-ViT-B/16 in our case. The semantic distance between the original and corresponding synthetic images was evaluated by extracting features through Visual DNA  and comparing their distribution, image-wise or dataset-wise, using the Earth Mover's Distance (EMD).  At the image level, the distance between each anonymized-real image pair was calculated and then the mean and standard deviation were calculated on the anonymized dataset. Results for data set-level visual DNA distance are available in the Supplementary material. Fréchet Inception Distance (FID) \cite{DBLP:journals/corr/HeuselRUNKH17} was also used to measure the overall quality of the synthetic dataset.   

\paragraph{Anonymization evaluation}

The risk of re-identification was measured by comparing each anonymized image against the real dataset and retrieving the K-Nearest Neighbors (K-NN). Unlike previous works, \textit{we consider both face-level and image-level re-identification protocols}. 

The Face-level protocol mimics a Facial Recognition system, as in previous works \cite{barattin2023attribute}. The largest face-crop was extracted using the MTCNN \cite{Zhang2016JointFD} detector, and face embeddings are extracted from crop using a FaceNet model \cite{Schroff2015FaceNetAU} pre-trained on VGGFace2 \cite{Cao2017VGGFace2AD} or CASIA WebFace \cite{Yi2014LearningFR}. The Image-level protocol, instead, simulates the use of a traditional web-based image retrieval engine and thus takes the background into account. It does so by extracting the embeddings from the entire image using the CLIP visual encoder to perform the K-NN search.  {In both cases, the metrics used were re-identification rate at different ranks (Re-ID@K) for K=1, 5 and 10 and mean Average Precision (mAP).} For the face-level protocol, the Re-ID@1 corresponds to the ``re-identification rate'' defined by the authors of FALCO \cite{barattin2023attribute}. 
Given the small size of the datasets (e.g., 30k for CelebA-HQ ), we also tested all techniques using an image search engine (Google Lens \cite{googlelens}). The re-ID rate was calculated after filtering out images not indexed by the search engine.

\paragraph{Downstream tasks} To assess the reusability and utility of the generated images, we compared performance in various tasks against the original counterparts, including facial attribute prediction, valence/arousal prediction, emotion recognition, ethnicity classification, and caption generation. We also investigated the performance of models trained on anonymized images, particularly in facial attribute prediction, akin to the methodology employed by FALCO's authors \cite{barattin2023attribute}.  

%% file: sec/5_results.tex
\section{Results}
\label{sec:results}
\paragraph{Anonymization performance}

The re-identification rate (Re-ID@1)  for different methods are compared in \cref{tab:celeba-table} on CelebA-HQ and LFW, respectively. The Re-ID rate at different ranks and the mAP can be found in the Supplementary Material. Under the conventional face-level re-identification setting, our proposed models have slightly higher Re-ID@1 rates (CelebA-HQ: 1.8\%-1.9\%) compared to DP2 (CelebA-HQ: 0.8\%) and FALCO (CelebA-HQ: 0.2\%-0.4\%). Indeed, this metric favors methods focusing solely on anonymizing facial features; this trend is further confirmed by the fact that \IName{}, which explicitly conditions on facial features, outperforms \BName{} when no negative control is used.

Notably, when testing CLIP-based image-level re-identification, the dynamics shift. Particularly in LFW, in which faces cover a smaller portion of the overall image area wrt CelebA-HQ (21\% vs. 40\%), the performance of inpainting-based methods like DP2 is severely degraded (CelebA-HQ: 30.6\%, LFW: 81\%). On the other hand, we attribute FALCO’s low Re-ID@1 (CelebA-HQ: 10.9\%, LFW: 4.8\%) to the deliberate suppression of background information, as will be discussed later. \BName{} (CelebA-HQ: 8.7\%, LFW: 40\%) and \IName{} (CelebA-HQ: 18\%, LFW: 66.8\%) achieve intermediate results underlying the challenge of reproducing key features of the background, whilst preventing re-identification. 
Finally, we evaluated ReID using an image search engine (Google Lens\cite{googlelens}) to assess a realistic context in which the number of indexed images is much higher. The results show the main limitations of DP2, that as the area occupied by faces decreases, ReID increases, but highlight the promise of \BName{} in protecting from online searches (CelebA-HQ: 0.9\%, LFW: 0\%).

\begin{table}[tb]
\centering
\resizebox{\textwidth}{!}{%
\begin{tabular}{c|cccc|cccc}
  \multicolumn{1}{l|}{ } &
  \multicolumn{4}{c|}{\textbf{CelebA-HQ}}&
  \multicolumn{4}{c}{\textbf{LFW}}\\ \hline
  \multicolumn{1}{c|}{\textbf{\textbf{Encoding}}} &
  \multicolumn{1}{c|}{\textbf{VGGFace2}} &
  \multicolumn{1}{c|}{\textbf{CASIA}} &
  \multicolumn{1}{c|}{\textbf{CLIP-B}}  & 
  \multicolumn{1}{c|}{\textbf{LENS}}&
  \multicolumn{1}{c|}{\textbf{VGGFace2}} &
  \multicolumn{1}{c|}{\textbf{CASIA}} &
  \multicolumn{1}{c|}{\textbf{CLIP-B}}  &
  \multicolumn{1}{c}{\textbf{LENS}}\\ \hline
  \multicolumn{1}{c|}{\textbf{DeepPrivacy2\cite{hukkelaas2023deepprivacy2}}} &
  \multicolumn{1}{c|}{\underline{0.008}} &
  \multicolumn{1}{c|}{\underline{0.008}} &
  \multicolumn{1}{c|}{0.306}  &
  \multicolumn{1}{c|}{0.080}&
  \multicolumn{1}{c|}{0.023} &
  \multicolumn{1}{c|}{\underline{0.017}} &
  \multicolumn{1}{c|}{0.810}  &
  \multicolumn{1}{c}{0.196}\\
  \multicolumn{1}{c|}{\textbf{FALCO\cite{barattin2023attribute}}} &
  \multicolumn{1}{c|}{\textbf{0.002}} &
  \multicolumn{1}{c|}{\textbf{0.004}} &
  \multicolumn{1}{c|}{\underline{0.109}}  &
  \multicolumn{1}{c|}{\textbf{0.002}} &
  \multicolumn{1}{c|}{\textbf{0.009}} &
  \multicolumn{1}{c|}{\textbf{0.010}} &
  \multicolumn{1}{c|}{\textbf{0.048}}  &
  \multicolumn{1}{c}{-} \\
  \multicolumn{1}{c|}{\textbf{Ours-Base} ($as=1.0$)} &
  \multicolumn{1}{c|}{0.096} &
  \multicolumn{1}{c|}{0.100} &
  \multicolumn{1}{c|}{0.211}  &
  \multicolumn{1}{c|}{0.015} &
  \multicolumn{1}{c|}{0.102} &
  \multicolumn{1}{c|}{0.116} &
  \multicolumn{1}{c|}{0.391}  &
  \multicolumn{1}{c}{\underline{0.025}}\\
  \multicolumn{1}{c|}{\textbf{Ours-Base} ($as=1.25$)} &
  \multicolumn{1}{c|}{0.018} &
  \multicolumn{1}{c|}{0.019} &
  \multicolumn{1}{c|}{\textbf{0.087}}  &
  \multicolumn{1}{c|}{\underline{0.009}} &
  \multicolumn{1}{c|}{\underline{0.021}} &
  \multicolumn{1}{c|}{0.023} &
  \multicolumn{1}{c|}{\underline{0.189}}  &
  \multicolumn{1}{c}{\textbf{0}} \\
  \multicolumn{1}{c|}{\textbf{Ours-Light}} &
  \multicolumn{1}{c|}{0.046} &
  \multicolumn{1}{c|}{0.036} &
  \multicolumn{1}{c|}{0.399}  &
  \multicolumn{1}{c|}{0.041}&
  \multicolumn{1}{c|}{0.084} &
  \multicolumn{1}{c|}{0.067} &
  \multicolumn{1}{c|}{0.668}  &
  \multicolumn{1}{c}{\underline{0.025}}\\ \hline
\end{tabular}%
}
\caption{Re-identification rate for CelebA-HQ \cite{karras2017progressive} and LFW \cite{huang2008labeled}.}
\label{tab:celeba-table}
\end{table}

\begin{figure}[tb]
    \centering
    \includegraphics[width=\linewidth]{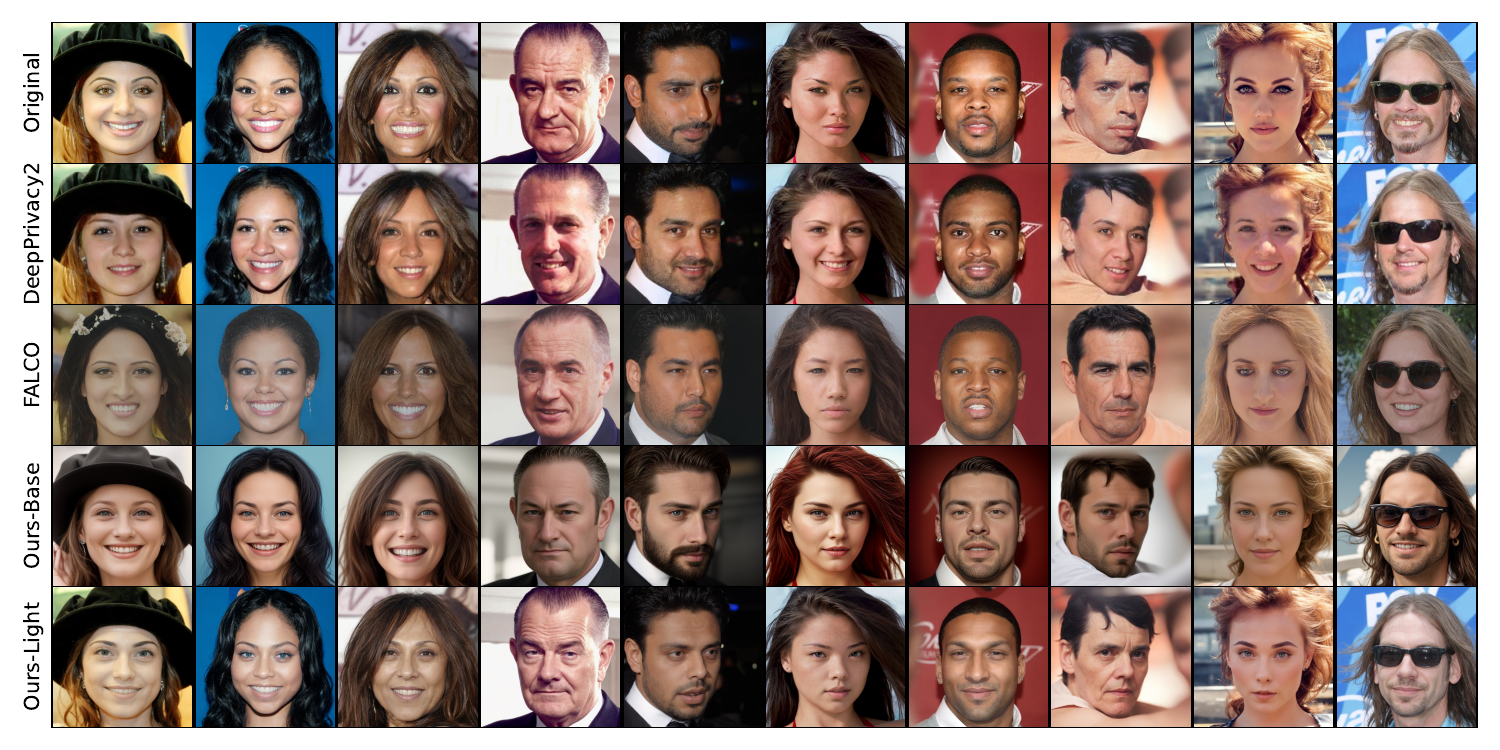}
    \caption{Anonymization results on CelebA-HQ \cite{karras2017progressive} in comparison with DeepPrivacy2 (DP2) \cite{hukkelas23DP2} and FALCO \cite{barattin2023attribute}.}
    \label{fig:samples_celeb}
\end{figure}

\begin{figure}[tb]
    \centering
    \includegraphics[width=\linewidth]{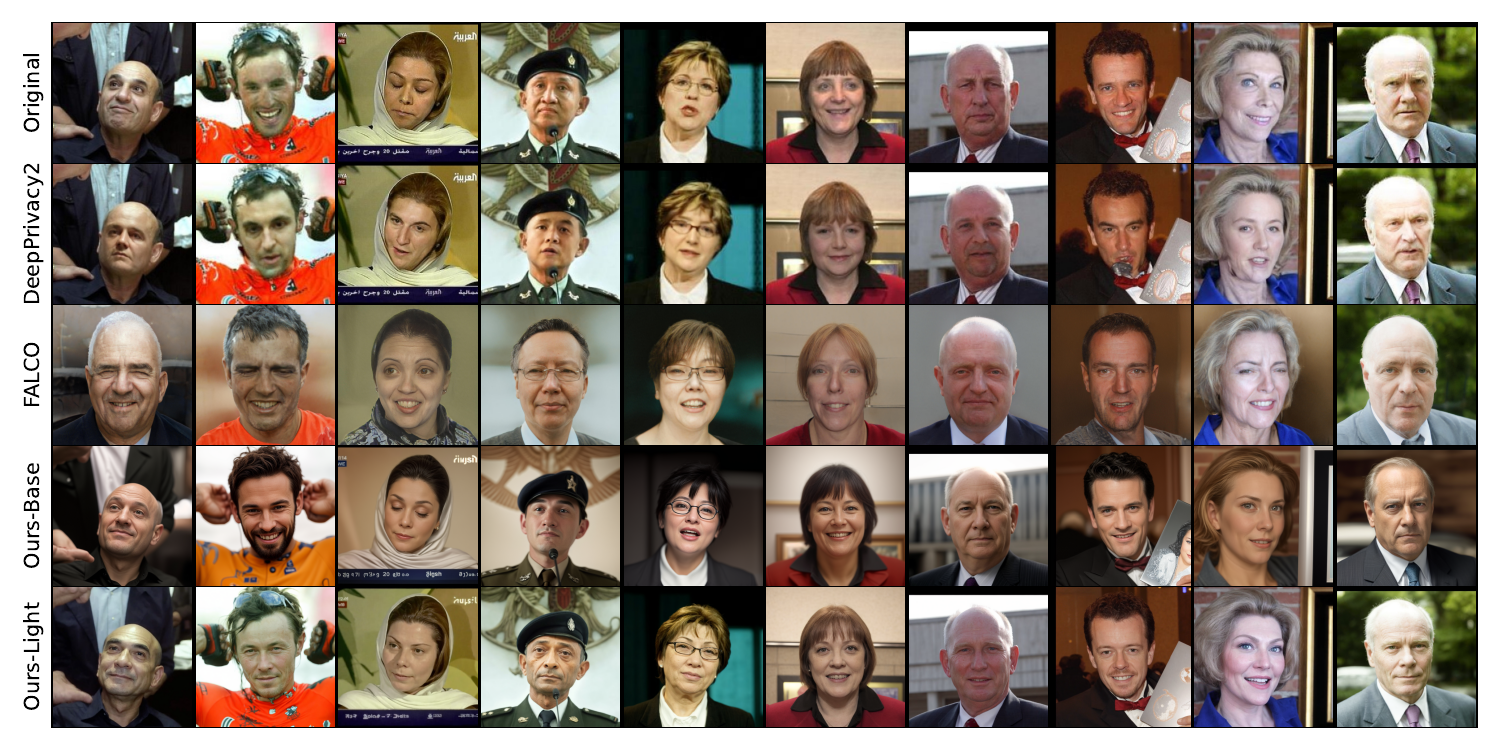}
    \caption{Anonymization results on LFW \cite{huang2008labeled} in comparison with DeepPrivacy2 (DP2) \cite{hukkelas23DP2} and FALCO \cite{barattin2023attribute}.}
    \label{fig:samples_lfw}
\end{figure}

\begin{figure}[tb]
    \centering
    \includegraphics[width=\linewidth]{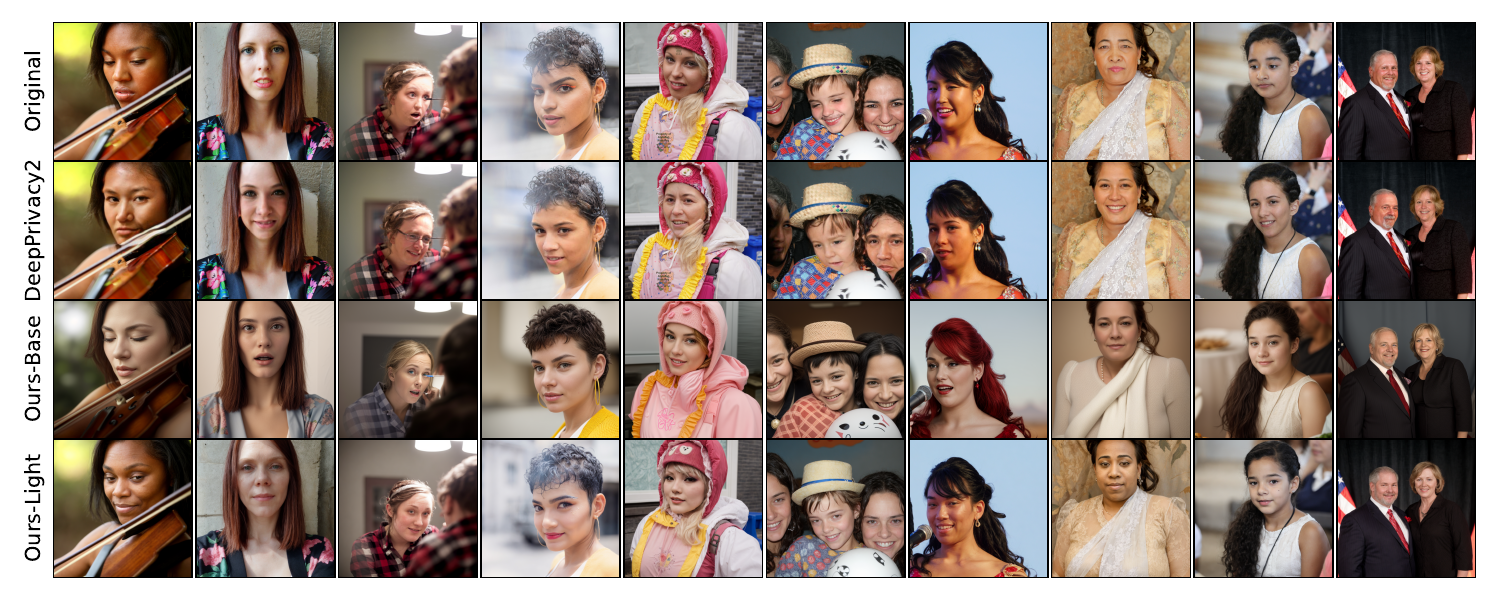}
    \caption{Anonymization results on FFHQ-in-the-wild \cite{karras2019style} in comparison with DeepPrivacy2 (DP2) \cite{hukkelas23DP2}.}
    \label{fig:samples_ffhq}
\end{figure}

\paragraph{Qualitative image quality assessment}
In \cref{fig:samples_celeb,fig:samples_lfw,fig:samples_ffhq} randomly selected examples of anonymized images, including both portraits and scenes, are shown. The proposed LDMs generate high-quality images; they retain the facial attributes of the original subject to a much greater extent than DP2 and, compared to FALCO, preserve the overall structure and background characteristics.

Compared to \BName{} and FALCO, \IName{} mitigates palette shifts and generate images with brighter colors, and in general closer to the original image. The embedding of facial attributes in \IName{} more accurately preserves key facial attributes, and is generally better at preserving the original ethnicity, a crucial issue for \BName{}. However, the latter excels at retaining certain small features, such as glasses,  occasionally lost during the face-swapping process in \IName{}.  FALCO, while generating highly realistic facial images, tends to produce darker palettes, and typically destroys important information, such as the background and clothing style, that could be relevant for downstream tasks. Additionally, it lacks the ability to anonymize complex scenes with multiple persons, a shortcoming that is successfully addressed by our method, as shown in \cref{fig:samples_ffhq}. More qualitative results can be found in the Supplementary Material.

\begin{table}[tb]
\center
\resizebox{0.7\columnwidth}{!}{%
\begin{tabular}{c|c |c |c |c}

\hline
\textbf{Method} & \textbf{FFHQ in-the-wild} & \textbf{WIDER FACE} & \textbf{CelebA-HQ} & \textbf{LFW} \\ \hline
&\multicolumn{4}{c}{FID}\\
\hline
\textbf{DeepPrivacy2\cite{hukkelaas2023deepprivacy2}} & \textbf{20.7} & \textbf{12.9} & 49.4 & \textbf{8.9} \\
\textbf{FALCO\cite{barattin2023attribute}} & - & - & 41.2 & 96.7 \\
\textbf{Ours-Base} ($as=1.0$) & 67.8 & 84.7 & \underline{35.4} & 62.8 \\
\textbf{Ours-Base} ($as=1.25$) & 76.5 & 94.2 & 41.5 & 78.2 \\
\textbf{Ours-Light} & \underline{60.1} & \underline{81.8} & \textbf{28.7} & \underline{54.6} \\
\hline
&\multicolumn{4}{c}{Visual DNA}  \\ \hline
\textbf{DeepPrivacy2\cite{hukkelaas2023deepprivacy2}} & \textbf{5.2}\(\pm\)\textbf{1.8} & \textbf{3.2}\(\pm\)\textbf{2.6} & \textbf{10.4}\(\pm\)\textbf{1.3} & \textbf{5.6}\(\pm\)\textbf{1.0} \\
\textbf{FALCO\cite{barattin2023attribute}} & - & - & 15.2\(\pm\)2.1 & 20.1\(\pm\)1.9 \\
\textbf{Ours-Base} ($as=1.0$) & 14.3\(\pm\)2.2 & 16.0\(\pm\)2.5 & 15.9\(\pm\)2.0 & 17.8\(\pm\)2.1 \\
\textbf{Ours-Base} ($as=1.25$) & 16.0\(\pm\)2.3 & 17.7\(\pm\)2.7 & 17.7\(\pm\)2.1 & 20.0\(\pm\)2.1 \\
\textbf{Ours-Light} & \underline{11.3\(\pm\)1.2} & \underline{12.8\(\pm\)1.9} & \underline{12.3\(\pm\)1.4} & \underline{14.0\(\pm\)1.5} \\
\hline
\end{tabular}%
}
\caption{Image quality results on FFHQ in-the-wild \cite{karras2019style}, WIDER FACE \cite{yang2016wider}, CelebA-HQ \cite{karras2017progressive} and LFW \cite{huang2008labeled}. FID distances are calculated between the real and anonymized datasets; Visual DNA distance is calculated on each pair of real-anonymized image, and then mean and standard deviation are calculated on the entire dataset. }
\label{tab:FID_pairs}
\end{table}

\paragraph{Quantitative image quality assessment}
Results from FID and Visual DNA are presented in \cref{tab:FID_pairs}. Results are calculated on a sample of 400 images from the validation sets of FFHQ in-the-wild and WIDER FACES each, and on CelebA-HQ and LFW. \CName{} achieves lower FID scores than  FALCO and DP2 on portrait datasets (LFW and CelebA-HQ). We argue this result is due to the superior ability to generate high-resolution images with low artifacts. Inpainting methods like DP2 occasionally introduce artifacts, whereas images generated by FALCO are highly realistic but with lower resemblance to the original. DP2 achieves very low FID in datasets featuring predominantly scenes (FFHQ-in-the-wild and WIDER), as expected given that the method modifies only a minor part of the image; however, as a result the original image can be recovered from the background, as determined in  \cref{tab:celeba-table}. The distribution of the Visual DNA distance further corroborate these results. Unlike FID, this metric measures to what extent each pair of real-anonymized images are consistent. From this perspective, \IName{} outperforms FALCO, which in turn performs similarly to \BName{}. This underscores the effectiveness of our approach in preserving the semantic content of individual images.

\paragraph{Downstream tasks}
Performance on downstream tasks is reported in \cref{tab:down_task}. Classifiers trained on images generated by \IName{} closely approach the performance levels wrt training on the original images (AUC: 0.890 vs. 0.904 for inner attributes; 0.919 vs. 0.922 for outer attributes). DP2 achieves good performance on caption generation and outer facial attribute prediction that rely heavily on content unchanged by inpainting. Unlike FALCO, our method achieves strong performance for both inner and outer face attributes.  The lower performance of DP2 and FALCO on ethnicity classification raises an important question for future work: how much does (not) preserving ethnicity influence the performance obtained at ReID?
This comprehensive evaluation highlights the robustness and efficacy of our method across a spectrum of downstream tasks, establishing its potential for practical applications.

\begin{table*}[tb]
\centering
\resizebox{\textwidth}{!}{%
\begin{tabular}{c|cc|cccccc}
\hline
\textbf{Downstream task} &
  \multicolumn{2}{c|}{\textbf{Anonymized $\rightarrow$ Real }} &
  \multicolumn{6}{c}{\textbf{Real $\rightarrow$ Anonymized }}\\ \hline
\textbf{Method} & 
\textbf{Inner face attrs} &
\textbf{Outer face attrs} &
\textbf{Inner face attrs} & 
\textbf{Outer face attrs}&
\textbf{Emotions} &
\textbf{Valence-Arousal} &
\textbf{Caption} &
\textbf{Ethnicity}\\
 & (AUC ↑) & (AUC ↑) & (AUC ↑) & (AUC ↑) & (AUC ↑) & (MSE ↓) & (CLIP Score ↑)  & (Accuracy ↑)\\
  \hline
\textbf{Original}       & 0.904& 0.922
& -& -& -& -& -
&-\\
\textbf{DP2} & 0.866& \underline{0.914}
& 0.823& 0.976& 0.664& 0.18& \textbf{0.83}&0.674\\
\textbf{FALCO}          & 0.883& 0.902
& \underline{0.898}& 0.944& 0.777& 0.08& 0.62
&0.654\\
\textbf{Ours-Base} ($as=1.0$)& \underline{0.887}& 0.906
& \textbf{0.921}& \underline{0.978}& \textbf{0.927}& \textbf{0.05}& \textbf{0.83}
& \textbf{0.777}\\
\textbf{Ours-Base} ($as=1.25$)& 0.874& 0.896
& 0.889& 0.962& \underline{0.903}& \underline{0.06}& \underline{0.81}
& 0.744\\
\textbf{Ours-Light}& \textbf{0.890}& \textbf{0.919}& 0.874& \textbf{0.980}& 0.776& 0.09& 0.76&\underline{0.758}\\ \hline
\end{tabular}%
}
\caption{Performance on downstream tasks. All tasks are evaluated on CelebA-HQ except caption generation (LFW). For the Anonymized $\rightarrow$ Real setting, we finetune a pretrained model on 29K images and test on the remaining 1K. }
\label{tab:down_task}
\end{table*}

\paragraph{Inference time}

\IName{} achieves a roughly $3.8\times$ speed up over \BName{} (2.8 s/image vs. 10.8 s/image at $768^2$ resolution). Additional details are reported in the Supplementary Material. 

%% file: sec/6_conclusion.tex
\section{Conclusions}
\label{sec:conclusion}

In this paper, we presented \CName{}, a novel anonymization framework based on LDMs. \BName{} exploits a combination of pre-trained ControlNets and introduces an anonymizazion guidance based on the original image, while \IName{} trains a lightweight IP-Adapter to encode key elements of the scene and facial attributes of each person. The former achieves stronger anonymization, while the latter generally preserves image content better and simultaneously reduces the inference time by 75\%. 
Compared to state-of-the-art, we anonymize complex scenes by introducing variations in the faces, bodies, and background elements.  Future work will aim at further improving the delicate trade-off between preserving the semantic content, while ensuring effective de-identification. In addition, we plan to delve further into the evaluation, investigating how effectively properties of the original image are preserved, from the standpoints of human perception as well as of selected downstream tasks. 

%% file: sec/7_ack.tex
\section{Acknowledgment}
\label{sec:ack}
This work was supported by the EU H2020 AI4Media No. 951911 project.